\documentclass[letterpaper, 10 pt, conference]{ieeeconf}  

\IEEEoverridecommandlockouts                              
\usepackage[utf8]{inputenc}
\usepackage[T1]{fontenc}

\usepackage{float} 

\usepackage[english]{babel}
\usepackage{amsmath,amsfonts,amssymb}
\usepackage{pdfpages}
\usepackage{setspace}
\usepackage{titlesec}

\usepackage{algorithm}
\usepackage{algpseudocode}
\usepackage{graphicx}
\usepackage{tikz}
\usetikzlibrary{arrows.meta,positioning,shapes.geometric,decorations}
\usepackage[hidelinks]{hyperref} 

\title{\LARGE \bf
Efficient Online Learning with Predictive Coding Networks:
Exploiting Temporal Correlations \thanks{\tt{Accepted at EdgeAI4R Workshop, IEEE/RSJ International Conference on Intelligent Robots and Systems (IROS) 2025}}
}

\author{%
Darius Masoum Zadeh-Jousdani$^{*}$,
Elvin Hajizada$^{*\dagger}$ and
Eyke Hüllermeier \\[0.8em]
\textit{Institute for Informatics, University of Munich (LMU), Munich, Germany}\\[0.5em]
{\small $^{*}$These authors contributed equally to this work.}\\
{\small $^{\dagger}$Corresponding author email: hajizada.elvin@campus.lmu.de}
}

\begin{document}

\maketitle

\thispagestyle{empty}
\pagestyle{empty}

\begin{abstract}
Robotic systems operating at the edge require efficient online learning algorithms that can continuously adapt to changing environments while processing streaming sensory data. Traditional backpropagation, while effective, conflicts with biological plausibility principles and may be suboptimal for continuous adaptation scenarios. The Predictive Coding (PC) framework offers a biologically plausible alternative with local, Hebbian-like update rules, making it suitable for neuromorphic hardware implementation. However, PC's main limitation is its computational overhead due to multiple inference iterations during training. We present Predictive Coding Network with Temporal Amortization (PCN-TA), which preserves latent states across temporal frames. By leveraging temporal correlations, PCN-TA significantly reduces computational demands while maintaining learning performance. Our experiments on the COIL-20 robotic perception dataset demonstrate that PCN-TA achieves 10\% fewer weight updates compared to backpropagation and requires 50\% fewer inference steps than baseline PC networks. These efficiency gains directly translate to reduced computational overhead for moving another step toward edge deployment and real-time adaptation support in resource-constrained robotic systems. The biologically-inspired nature of our approach also makes it a promising candidate for future neuromorphic hardware implementations, enabling efficient online learning at the edge.


\end{abstract}


\section{INTRODUCTION}

Backpropagation is a widely used and highly efficient learning algorithm. However, the prevailing consensus today is that the brain is unlikely to implement backpropagation in its exact form \cite{lillicrap2020backpropagation}. Most theories of learning in the brain postulate that learning occurs purely locally between neurons, without requiring distant error signals. This type of learning algorithm is also known as Hebbian plasticity, often summarized by the famous phrase: "Cells that fire together wire together" \cite{hebb2005organization}. Hence, reliance on global error information makes backpropagation biologically implausible. To achieve a biologically plausible learning algorithm, we must seek alternatives that adhere to local learning rules.

Predictive Coding is a theory of how the brain learns. The fundamental difference between Predictive Coding and backpropagation lies in how they propagate and use error signals in a neural network. Predictive coding theory emphasizes that each neuron interacts only with nearby neurons, typically those in adjacent layers, such as its inputs and outputs \cite{rao1999predictive}. This stands in contrast to backpropagation, which requires each unit to access information that depends on the entire network structure, as it computes gradients using the chain rule, which requires knowledge of the entire pathway from output to input. A core principle of Predictive Coding is that each layer in a hierarchical neural network generates predictions about the activity of the layer below. learning is guided by the minimization of local prediction errors, calculated as the difference between the actual activity of lower-layer neurons and the top-down predictions. This process follows the principle of learning through error minimization, whereby the network continuously updates its internal representations and synaptic weights to reduce these discrepancies over time. A distinctive feature of Predictive Coding is that, rather than immediately updating weights for each new input, the network first undergoes an inference phase. During this phase, it iteratively adjusts its internal or hidden states in order to minimize prediction error. In essence, the network attempts to ”explain” the current input by refining its internal model, aligning top-down predictions as closely as possible with the actual neuronal activity. This iterative process continues until the prediction errors converge to a minimum, allowing the network to make more informed and plausible updates.

Millidge et al. ~\cite{PredictiveCodingBackprop} introduced a practical implementation of the predictive coding theory, which we will hereafter refer to simply as PCN. This framework differs from classical predictive coding theory, as it operates under the fixed-prediction assumption: during the initial feedforward phase, the input propagates through each layer of the network until it reaches the output layer. The activations of neurons are used as a prediction, which remains the same throughout the entire inference phase of the given sample. In the inference phase, based on label, input, and the predictions, each neuron's internal state is iteratively updated to minimize the local prediction error. Once the number of predefined inference steps are finished, the leftover error are further minimized by a single weight update. Thanks to the fixed-prediction assumption, this PCN approximates backpropagation.

However, Millidge et al. argued in \cite{millidge2022predictive} that the main drawback of his PC version lies in its computational cost: unlike Backpropagation, which computes gradients in a single pass, this PC variant requires iterative inference with repeated updates until prediction errors converge, performed independently for each input sample. Nevertheless, in online learning scenarios such as video streams, latent states do not need to be reinitialized from scratch for every frame, since consecutive inputs are temporally correlated. To exploit this property we introduce a \textbf{Predictive Coding Network with Temporal Amortization (PCN-TA)}, a variant of the baseline PCN that maintains latent states across frames, effectively leveraging temporal continuity in sequential data. This design should reduce the number of required inference iterations and the number of weight updates. Before presenting the methodology of our experiments, we first describe the PCN-TA in greater detail, followed by a discussion of our experimental methods and results and concluding with a brief discussion and summary.\\

\section{Mathematical Formulations of PCN-TA}
For both the PCN-TA and the PCN, we have an arbitrary computation graph, where we consider $\mathcal{G} = \{\mathbb{Y}, \mathbb{V}, \mathbb{E}\}$. Each edge $y_i \in \mathbb{Y}$ corresponds to the application of an activation function that computes a vertex $v_i \in \mathbb{V}$. Additionally, we define error units $\epsilon_i \in \mathbb{E}$ as the difference between the actual state value $v_i$ and the top-down prediction $\hat{v}_i$: $\epsilon_i = v_i - \hat{v}_i$. In a Predictive Coding network, each hierarchical layer is composed of functional units that include a \textbf{prediction error neuron} $\epsilon_i$ and its associated \textbf{state neuron} $v_i$, representing the actual state of the corresponding vertex.\\

In contrast to Backpropagation, which computes the gradient of the global loss, Predictive Coding computes and minimizes the gradient of the variational free energy (VFE) $\mathcal{F}$. The VFE, an objective from variational inference that renders Bayesian learning tractable, serves as the function that guides updates of the state neurons $v_i$. For a more detailed explanation of the VFE we refer the reader to \cite{bogacz2017tutorial} and \cite{millidge2021predictive}.\\

In PCN-TA, the initialization for the first frame follows the standard Predictive Coding procedure. Specifically, during the initial feedforward phase, the input is propagated through the network as in ANNs, where the activations are used as predictions $\hat{v}_i$. These predictions are kept same throughout the inference phase for the given sample. This is the fixed-prediction assumption and needed to approximate backpropagation. Additionally initial states ($v_i$s) of the neurons are also set to the predicted values ($\hat{v}_i$). The next steps of the inference phase consists of iterative update of these state values, according to $v_i^{t+1} = v_i^t + \eta_v \tfrac{\partial \mathcal{F}}{\partial v_i^t}$, with $\eta_v$ denoting the inference learning rate, which serves as the step size in the optimization process, and $\tfrac{\partial \mathcal{F}}{\partial v_i^t}$ representing the gradient of the VFE $\mathcal{F}$. The backward phase is repeated iteratively until the gradient of the variational free energy with respect to each layer \(v_i\) converges to an equilibrium point, that is, when \(\partial \mathcal{F} / \partial v_i\) = 0. After completing inference for the first sample, we store the resulting hidden state values to be reused in the next frame $x$. For full mathematical description of this process, refer to the pseudocode in Algorithm~\ref{alg:gpc1}.

\begin{algorithm}[H]
\caption{PCN-TA for first sample}
\label{alg:gpc1}
\begin{algorithmic}[1]
\State \textbf{Input:} Dataset $\mathcal{D}$; first sample $x^1 \in \mathcal{D}$; label $\mathcal{L}$; $\mathbf{L}\}$; Inference learning rate $\eta_v$; Weight learning rate $\eta_\theta$
\For {first sample x in dataset}
    \State Let $(x^1, L)$ be the first sample with its label
    \State $v_0 \gets x^1$
    
    \Comment Forward pass to compute predictions
    \ForAll{$\hat{v}_i \in \mathbb{V}$}
        \State $\hat{v}_i \gets f(\mathcal{P}(\hat{v}_i); \theta)$
    \EndFor

    \State $\epsilon_L \gets L - \hat{v}_L$

    \Comment Backward phase: state updates via free energy descent
    \While{not converged}
        \ForAll{$(v_i, \epsilon_i) \in \hat{\mathcal{G}}$}
            \State $\epsilon_i \gets v_i - \hat{v}_i$
            \State $v_i^{t+1} \gets v_i^t + \eta_v \frac{\partial \mathcal{F}}{\partial v_i^t}$
        \EndFor
    \EndWhile

    \Comment Update weights after convergence
    \ForAll{$\theta_i \in \mathbb{E}$}
        \State $\theta_i^{t+1} \gets \theta_i^t + \eta_\theta \frac{\partial \mathcal{F}}{\partial \theta_i^t}$
    \EndFor

    \Comment after inference phase, save the state values for the next sample
    
\EndFor
\end{algorithmic}
\end{algorithm}

When the inference phase begins for the next frame t, we first perform a standard feedforward pass from the input to the output, initializing the state values to their predicted counterparts. We then conduct a second feedforward pass, this time restoring all state values $v_i$ from the hidden states saved in the previous frame. This strategy streamlines the inference phase by reducing the number of required iterations: the network first generates predictions from input to output, after which the saved hidden states are reinstated, avoiding the need to reinitialize the inference process. From that point onward, the procedure mirrors that of the first sample, with iterations continuing until convergence. This process is repeated frame by frame, as illustrated in Algorithm~\ref{alg:gpc2}. For real code implementation see the following repository \cite{PCNTAREPO}\\

\begin{algorithm}[H]
\caption{Inference for Subsequent Samples ($\text{number} \ne 0$)}
\label{alg:gpc2}
\begin{algorithmic}[1]
\State \textbf{Input:} Dataset $\mathcal{D}$; sample $x \in \mathcal{D}$; label $\mathcal{L}$; $\mathbf{L}\}$; Inference learning rate $\eta_v$; Weight learning rate $\eta_\theta$
\For {sample $x$ in dataset}
    \State Let $(x, L)$ be the sample with its label
    \State $v_0 \gets x$
    
    \Comment first Forward pass to compute state values
    \ForAll{$\hat{v}_i \in \mathbb{V}$}
        \State $\hat{v}_i \gets f(\mathcal{P}(\hat{v}_i); \theta)$
    \EndFor

    \Comment Second feedforward pass, initializing predictions from the hidden states of the previous frame
    
    \ForAll{${v}^t_i \in \mathbb{V}$}
        \State ${v}^t_i \gets {v}^{t-1}_i$
    \EndFor

    \State $\epsilon_L \gets L - \hat{v}_L$
    
    \Comment The procedure then continues as described in Algorithm~\ref{alg:gpc2}

    \Comment Backward phase: state updates via free energy descent
    \While{not converged}
        \ForAll{$(v^t_i, \epsilon_i) \in \hat{\mathcal{G}}$}
            \State $\epsilon_i \gets v^t_i - \hat{v}_i$
            \State $v_i^{t+1} \gets v_i^t + \eta_v \frac{\partial \mathcal{F}}{\partial v_i^t}$
        \EndFor
    \EndWhile

    \Comment Update weights after convergence
    \ForAll{$\theta_i \in \mathbb{E}$}
        \State $\theta_i^{t+1} \gets \theta_i^t + \eta_\theta \frac{\partial \mathcal{F}}{\partial \theta_i^t}$
    \EndFor

    \Comment after inference phase, save the state values for the next sample
    
\EndFor
\end{algorithmic}
\end{algorithm}

For the further technical details of the previous work, we refer the reader to \cite{salvatori2023survey} and \cite{millidge2021predictive}, both of which provide a more in-depth exploration of the core principles of our Predictive Coding network and graph for a supervised learning scenario. 

\section{Methodology}
To demonstrate the advantages of our PCN-TA, we compared its accuracy against a standard PCN and a Backpropagation network, all implemented with the same CNN architecture and learning parameters. Training was conducted on the COIL-20 dataset~\cite{millidge2021predictive}, which is particularly well suited for our study because it provides temporally correlated data ideal for online learning in artificial neural networks. The models are trained on sequential video frames of 20 objects, showing how the objects change over time. In addition, we evaluate our model in a Class-Incremental Learning setting, where data are presented sequentially in batches of novel classes. The objective of these experiments is to show that PCN-TA achieves comparable accuracy with fewer inference iterations than a standard PCN, thereby reducing computational cost. Furthermore, we report the average number of weight updates between frames for the PCN-TA, the standard PCN, and the Backpropagation network, providing further evidence of the efficiency of our approach.\\

For both of these experiments, we had the same CNN architecture for all three models. Extensive hyperparameter tuning of the learning rate was conducted for all models, with optimal performance achieved at a weight learning rate of approximately 0.00004. The first layer in the ANN is a convolutional layer with an input size of (1, 128, 128), which fits grayscale images from the COIL-20 dataset with a resolution of 128x128 pixels. This convolutional layer has 124 filters with a kernel size of 5. The second layer is a pooling layer with a size of 2. The output of this layer is flattened and fed into a fully connected layer with 200 neurons. The data will be transformed to the next fully connected layer with 128 neurons. The output layer consists of 20 neurons, each representing an object in the COIL-20 dataset. The Mean Squared Error (MSE) metric will be used as our loss function. Our activation function for each layer, excluding the pooling, penultimate and output layers, is the ReLU function. For our penultimate layer, we will use a linear activation function. We used Ada as the optimizer. 

\section{Experiments}
We will first present the accuracy results of our image classification experiment, followed by an analysis of the average weight updates per frame across epochs for the PCN-TA, the standard PCN, and the Backpropagation trained network. 

\subsection{Accuracy results}
\begin{figure}[H]
  \centering
  \includegraphics[width=0.5\textwidth]{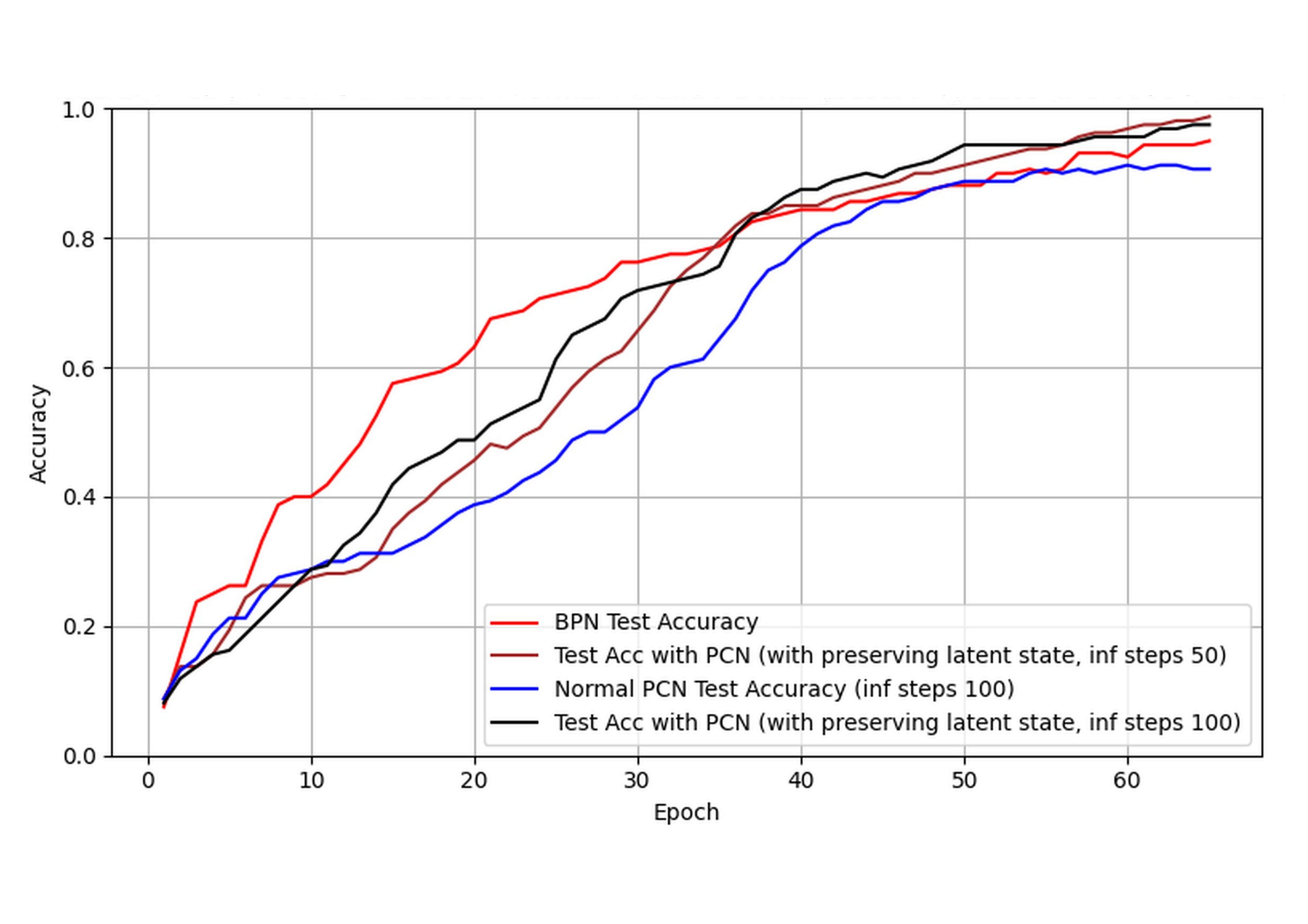} 
  \caption{The results of the image classification task show that backpropagation (red line) consistently achieves the highest average accuracy across all epochs. In contrast, the standard PCN with 100 inference iterations (blue line) performs the worst in nearly every epoch. Notably, the PCN-TA with only 50 inference iterations (brown line) outperforms the standard PCN, despite requiring fewer iterations.}
  \label{fig:accuracy results}
\end{figure}

Here we will show the experimental results of our image classification task and compare the accuracy of two PCN-TAs (with 50 and 100 inference iterations, respectively), a standard PCN with 100 iterations, and Backpropagation  (\autoref{fig:accuracy results}).\\

\textbf{Comparison between PCN-TAs and PCN:}
The main results are shown in Fig.~\ref{fig:accuracy results}. Notably the PCN-TA with 100 iterations has a better accuracy than the PCN with 100 inference iterations, but also the PCN-TA with only 50 iterations outperforms the standard PCN with 100 iterations, demonstrating the effectiveness of PCN-TA in exploiting temporally coherent data. This implies that the TA-PCN requires fewer inference steps to achieve comparable or even superior accuracy to a standard PCN. The reason is that it builds on the inference processes of previous frames, effectively making temporal predictions. By contrast, the  Predictive Coding network does not exploit prior inference and therefore requires a larger number of inference steps for each new frame. This also means that the PCN-TA needs fewer operations and is more efficient, possibly allowing for edge deployment. These findings underscore the substantial gains achieved when hidden states are preserved across frames, as opposed to reinitializing them from scratch for each new input.\\

\textbf{Comparison between PCN-TAs and backpropagation:} Both PCN-TAs (with 50 and 100 inference iterations) narrow the gap with backpropagation and ultimately surpass its performance in the later epochs. The PCN-TAs with 100 iterations approaches the accuracy of backpropagation most closely, while the 50-iteration PCN-TAs demonstrates an excellent trade-off between computational efficiency and performance.\\

\textbf{Comparison between  PCN and backpropagation:} The  PCN performs slightly worse than the backpropagation network by the end of training, though both appear to converge toward similar accuracy. This observation supports earlier findings~\cite{rosenbaum2022relationship}, which demonstrated that PCN is algorithmically equivalent to backpropagation, yielding similar gradient updates.

\subsection{Weight Update Sparsity}  
\begin{figure}[H]
  \centering
  \includegraphics[width=0.5\textwidth]{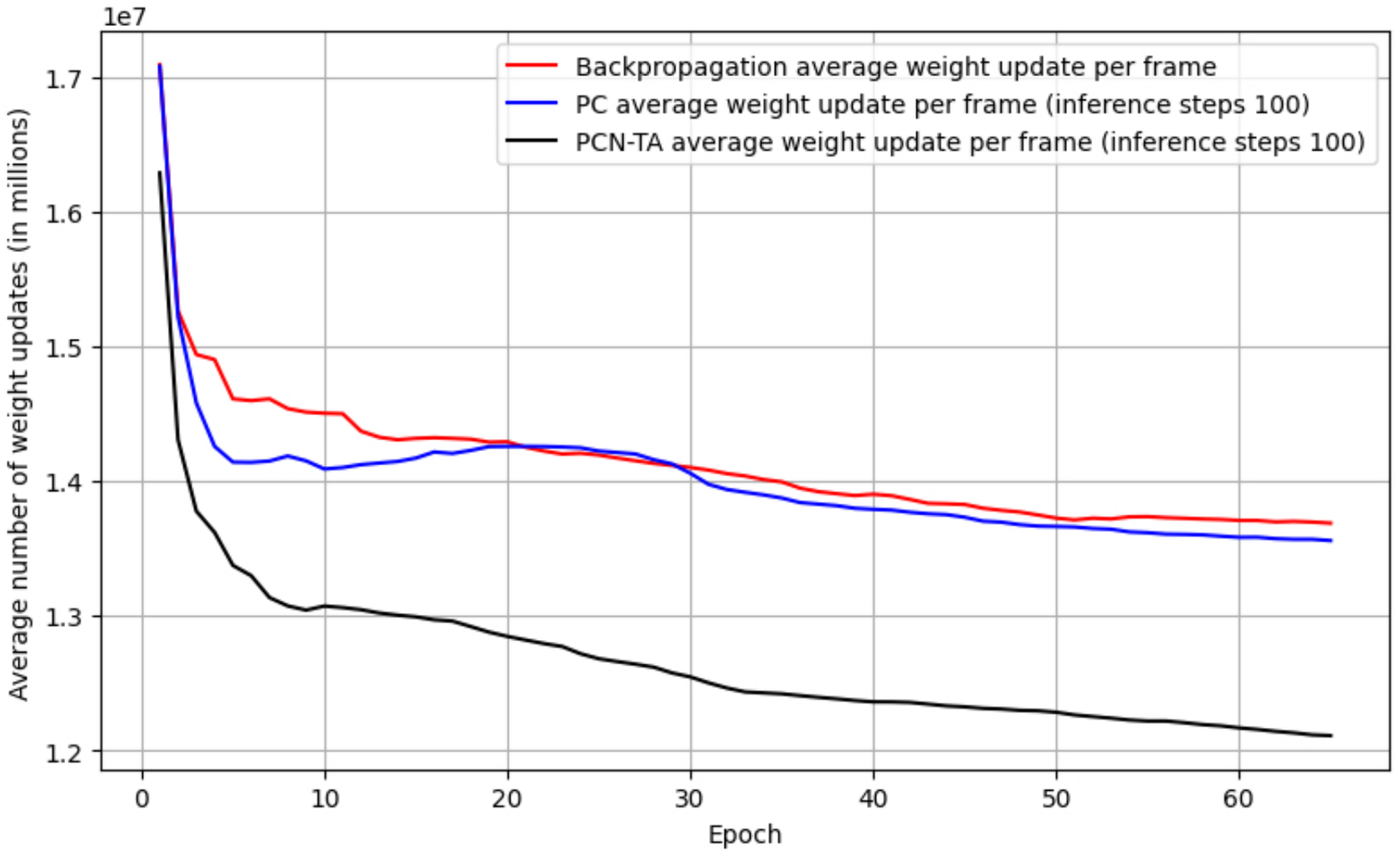} 
  \caption{This chart presents the average number of weight updates per frame across all epochs. The y-axis shows the average updates per frame, measured in millions. Backpropagation begins with nearly 1.7 million updates in the first epoch, with the  PCN exhibiting a similar value. In contrast, the PCN-TA starts with a considerably lower number of updates and maintains this advantage throughout training. Overall, the  PCN and backpropagation display comparable update counts, whereas the PCN-TA consistently achieves the lowest number across all epochs.}
  \label{fig:average update of weight updates per frame}
\end{figure}

In this experiment, we measured the average number of weight updates per frame during training for the PCN-TA with 100 inference iterations, the  PCN with 100 inference iterations, and a standard backpropagation network. Updates equal to zero were excluded from the count, and the values were averaged over the entire sample set \autoref{fig:average update of weight updates per frame}.\\

The results demonstrate that the PCN-TA requires significantly fewer weight updates per frame than the  PCN. It is important to note that PCN-TA departs from the goal of approximating Backpropagation, and hence its weight updates are not similar to the Backpropagation training. We think that, as only some parts of the images change from one frame to the next, the number of neurons that have a mismatch between prediction and value at the end of the inference phase is much fewer. By contrast, the  PCN, like backpropagation, must reinitialize for each new frame, resulting in a different or suboptimal convergence of state values at the end of the inference phase, hence triggering considerably more weight updates in the learning phase. This means that PCN-TA has a sparser weight updates.

\section{Discussion}
We demonstrated that for online learning from video streams, the PCN with a temporal amortization mechanism achieves higher test accuracy than the PCN, while requiring fewer inference steps and, as the weight update analysis shows, substantially fewer weight updates. This improvement arises from the preservation of latent (value) states across frames, which reduces the need for frequent synaptic weight adjustments: because consecutive frames are highly similar, the latent state from the previous frame provides a strong initial estimate for the next. By continuously maintaining and updating its latent state, the PCN-TA effectively anticipates incoming sensory input, in line with classical Predictive Coding theory. Unlike the  PCN, which reinitializes from scratch at each frame, the PCN-TA leverages temporal continuity to minimize redundant computation. Consequently, fewer inference steps are sufficient to achieve strong performance, as the model does not reset its internal state with every new input.

\section{Conclusion}
By incorporating Temporal Amortization, the PCN-TA exhibits greater sparsity in weight updates and achieves higher accuracy than both the  PCN and the backpropagation network. Nevertheless, the dataset used in this study was relatively small and lacked significant complexity, which
may limit the robustness and generalizability of the findings. Ideally, the model would have been trained
on a complete video sequence of an object and evaluated on a separate, distinct video sequence of the
same object captured under different conditions. The fixed-prediction assumption is also a limitation for PCN-TA, as it no longer tries to approximate backpropagation, but rather utilizes local sparse learning for more efficient training in online learning from temporally correlated data streams. Overall, we believe this is a valuable step towards edge-deployable learning systems. 

\bibliographystyle{plain}
\bibliography{main}

\end{document}